%% file: egpaper.tex
\documentclass[10pt,twocolumn,letterpaper]{article}

\usepackage{wacv}
\usepackage{graphicx}

\makeatletter
\@namedef{ver@everyshi.sty}{}
\makeatother

\usepackage{times}
\usepackage{epsfig}

\usepackage{amsmath}
\usepackage{amssymb}
\usepackage{booktabs}

\usepackage[table]{xcolor} 
\usepackage{tikz}
\usepackage{comment}
\usepackage{color}
\usepackage{multirow} 
\usepackage{siunitx}  %
\usepackage{subfig}
\usepackage{ulem} 
\usepackage{float}

\usepackage[symbol]{footmisc}

\def\wacvPaperID{673} 

\wacvalgorithmstrack   

\wacvfinalcopy 

\ifwacvfinal
\usepackage[breaklinks=true,bookmarks=false]{hyperref}
\else
\usepackage[pagebackref=true,breaklinks=true,colorlinks,bookmarks=false]{hyperref}
\fi

\input{commands}

\begin{document}

\title{SIRA: Relightable Avatars from a Single Image}

\author{Pol Caselles$^{1,2,3}$\\
\and
Eduard Ramon$^{1,2, \textbf{*}}$\\
\and
Jaime Garcia$^{1}$\\
\and
Xavier Giro-i-Nieto$^{2,3}$\thanks{This work was done prior to joining Amazon.}\\
\and
\vspace{-4mm}
Francesc Moreno-Noguer$^{3}$\\
\and
\vspace{-4mm}
Gil Triginer$^{1}$\\
\and
\vspace{-6mm}
\textit{\small$^1$Crisalix SA}
\quad \textit{\small$^2$Universitat Polit\`{e}cnica de Catalunya}
\quad \textit{\small$^3$Institut de Robòtica i Informàtica Industrial, CSIC-UPC}
\\
\vspace{-6mm}
}

\maketitle
\thispagestyle{empty}

\input{sections/0_abstract}

\input{sections/1_introduction}

\input{sections/2_related_work}
\input{sections/3_method}

\input{sections/4_experiments}

\input{sections/5_conclusions}

\clearpage
{\small
\bibliographystyle{ieee_fullname}
\bibliography{egbib}
}


\end{document}

%% file: commands.tex
\newcommand{\method}{SIRA}
\newcommand{\priorgeo}{SA-SM}
\newcommand{\priorcolor}{AF-SM}

\newcommand{\fsdf}{f^{\rm sdf}}

\newcommand{\ba}{\mathbf{a}}

\newcommand{\bc}{\mathbf{c}}

\newcommand{\bn}{\mathbf{n}}

\newcommand{\br}{\mathbf{r}}

\newcommand{\bv}{\mathbf{v}}

\newcommand{\bx}{\mathbf{x}}

\newcommand{\bz}{\mathbf{z}}

\newcommand{\bC}{\mathbf{C}}

\newcommand{\bI}{\mathbf{I}}

\newcommand{\bM}{\mathbf{M}}

\newcommand{\mC}{\mathcal{C}}

\newcommand{\mL}{\mathcal{L}}

\newcommand{\mP}{\mathcal{P}}

\newcommand{\mS}{\mathcal{S}}

\newcommand{\mbR}{\mathbb{R}} 
\newcommand{\mbS}{\mathbb{S}} 

\newcommand{\bdelta}{\boldsymbol{\delta}}

\newcommand{\bmu}{\boldsymbol{\mu}}

\newcommand{\bepsilon}{\boldsymbol{\epsilon}}
\newcommand{\bgamma}{\boldsymbol{\gamma}}

\newcommand{\btheta}{\boldsymbol{\theta}}

\newcommand{\bxi}{\boldsymbol{\xi}}

\newcommand{\bomega}{\boldsymbol{\omega}}

\newcommand{\argmin}{\operatornamewithlimits{arg\,min}}

%% file: sections/0_abstract.tex
\begin{abstract}
Recovering the geometry of a human head from a single image, while factorizing the materials and  illumination, is a severely ill-posed problem that requires prior information to be solved. Methods based on 3D Morphable Models (3DMM), and their combination with differentiable renderers, have shown promising results. However, the expressiveness of 3DMMs is limited, and they typically yield over-smoothed and identity-agnostic 3D shapes limited to the face region. Highly accurate full head reconstructions have recently been  obtained with neural fields that parameterize the geometry using multilayer perceptrons. The versatility of these representations has also proved effective for disentangling geometry, materials and lighting. However, these methods require several tens of input images. In this paper, we introduce \method, a method which, from a single image, reconstructs human head avatars with high fidelity geometry and factorized lights and surface materials. Our key ingredients are two data-driven statistical models based on neural fields that resolve the ambiguities of single-view 3D surface reconstruction and appearance factorization. Experiments show that \method{} obtains state of the art results in 3D head reconstruction while at the same time it successfully disentangles the global illumination, and the diffuse and specular albedos. Furthermore, our reconstructions are amenable to physically-based appearance editing and head model relighting.

\end{abstract}

%% file: sections/1_introduction.tex
\vspace{-4mm}
\section{Introduction}

Digitalizing humans into 3D relightable avatars is  key for a wide range of applications in e.g. augmented/virtual reality, 3D content production or the movie industry. In order to realistically render the captured models under new lighting conditions, it is not enough to just recover the 3D geometry, but also the rest of intrinsic properties of the scene need to be estimated, namely  surface materials and  scene illumination. It is specially challenging when the input data is acquired in non-controlled   conditions, and when the number of input images is small~\cite{wu2019mvf,ramon2019multi,bai2020deep,dou2018multi}. The single view setup is the most difficult scenario, and it poses a highly under-constrained problem that cannot be solved without a priori knowledge~\cite{tewari2017mofa,tuan2017regressing,richardson20163d,richardson2017learning,sela2017unrestricted,tran2018extreme}.

In order to recover the intrinsic properties of a scene from a single image, also known as inverse rendering, state of the art methods~\cite{tewari2017mofa,dib2021towards,smith2020morphable,lattas2020avatarme} introduce prior knowledge by means of 3D Morphable Models (3DMM)~\cite{bfm09,paysan20093d,li2017learning,tran2019towards,booth20163d,ploumpis2019combining,FLAME:SiggraphAsia2017}, which are combined  with deep neural networks. These models estimate the 3D geometry, spatially varying surface properties like the diffuse and specular albedos, as well as global illumination properties in the form of spherical harmonics or spherical gaussians. The regressed components are supervised using explicit ground truth data or in a self-supervised fashion in the image domain using differentiable renderers. However, the expressiveness of 3DMMs is limited, as they are biased towards low frequencies for both the geometry and albedo, and they are typically restricted to the facial area.

Recently, scene representation methods based on neural fields~\cite{xie2021neural} parameterized using multilayer perceptrons (MLP), have shown impressive results for the tasks of novel view synthesis and 3D reconstruction. One of the main advantages of neural fields in front of other representations like meshes or voxel grids is their great compromise between representational power and memory requirements. This, in combination with differentiable surface rendering~\cite{tewari2020state}, enables highly accurate 3D reconstructions and also to disentangle the scene into their intrinsic properties \cite{zhang2021physg}, leading to excellent models that can be rendered under novel lighting conditions. Their main limitation is that to supervise the learning process they require multiple views.

\begin{figure*}[t]
    \centering
    \includegraphics[width=1.0\textwidth]{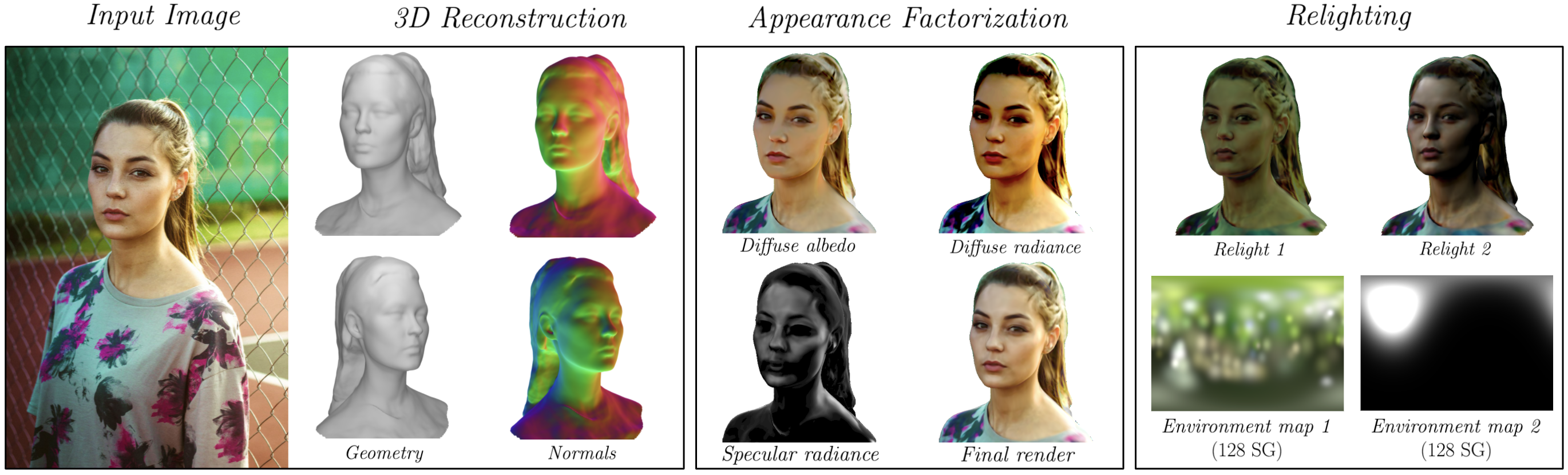}
    \caption{We introduce \method{}, a method which, from a single image, reconstructs human head avatars with high fidelity geometry and factorized lights and surface materials. This information can then be used for relighting purposes.}
    \vspace{-3mm}
    \label{fig:teaser}
\end{figure*}

To overcome these limitations we introduce \method{} (Fig. \ref{fig:teaser}), a neural field representation of the head, including hair and shoulders, that can be learned {\it from one single human portrait}. We approach the inverse rendering problem in two steps. First, we recover the 3D geometry by optimizing a signed distance function in combination with surface rendering, in the style of \cite{yariv2020multiview}. To resolve the inherent ambiguities of this problem, we introduce a novel shape and appearance statistical model (\priorgeo) that is used throughout the reconstruction process.  Then, in a second step where the 3D geometry is already estimated, we factorize the appearance into diffuse and specular albedos, and global illumination. For this,  we leverage an appearance factorization statistical model (\priorcolor), which is trained in a self-supervised fashion via physically-based rendering. 

A thorough evaluation demonstrates that the reconstructed geometry compares favorably to recent state-of-the-art methods, while in addition we also provide material and lighting parameters. The outcomes of \method{} enable, for the first time, the digitalization of human heads into relightable avatars from a single image. In summary, our contributions are threefold:

\vspace{-2mm}
\begin{itemize}
    \item A novel shape and appearance statistical model (\priorgeo) that allows  to recover the 3D geometry of portrait scenes from a single image.
    \item A novel appearance factorization statistical model (\priorcolor) that splits surface radiance fields into diffuse and specular albedos, and global illumination.
    \item A  methodology for training the aforementioned statistical models and  using them to create relightable avatars of 3D heads, including hair and shoulders.
\end{itemize}

%% file: sections/2_related_work.tex
\vspace{-4mm}
\section{Related work}

\noindent\textbf{Inverse rendering from a single portrait image. } Recovering the 3D head geometry from a single image is an ill-posed problem. Typically, 3DMMs in combination with deep neural networks are used to learn a mapping from an input image to a 3D geometry \cite{tuan2017regressing,richardson20163d,richardson2017learning,tran2018extreme}. In addition, some methods decompose the scene into diffuse albedo, and global illumination in the form of spherical harmonics \cite{tewari2017mofa,dib2021towards,smith2020morphable,lattas2020avatarme}. Recently, \cite{smith2020morphable} introduced a morphable model that disentangles the albedo into diffuse and specular components, enabling their factorization at test time \cite{dib2021towards}. In \cite{lattas2020avatarme}, the diffuse and specular reflectance is modelled using image-to-image translation networks in the texture space, which are learnt using ground truth data. While these methods enable the inverse rendering of portrait scenes, they present a number of limitations \cite{egger20203d}. First, 3DMM are usually restricted to the facial region \cite{bfm09,booth20163d}. Second, these methods only model low frequency geometry and further post-processing is usually required to obtain fine details \cite{richardson2017learning}. And third, the topology of 3DMMs is fixed, which limits the shapes these methods can represent \cite{ploumpis2019combining}. These are important limitations for reconstructing realistic avatars.

\begin{figure*}[t]
    \centering
    \includegraphics[width=1.0\textwidth]{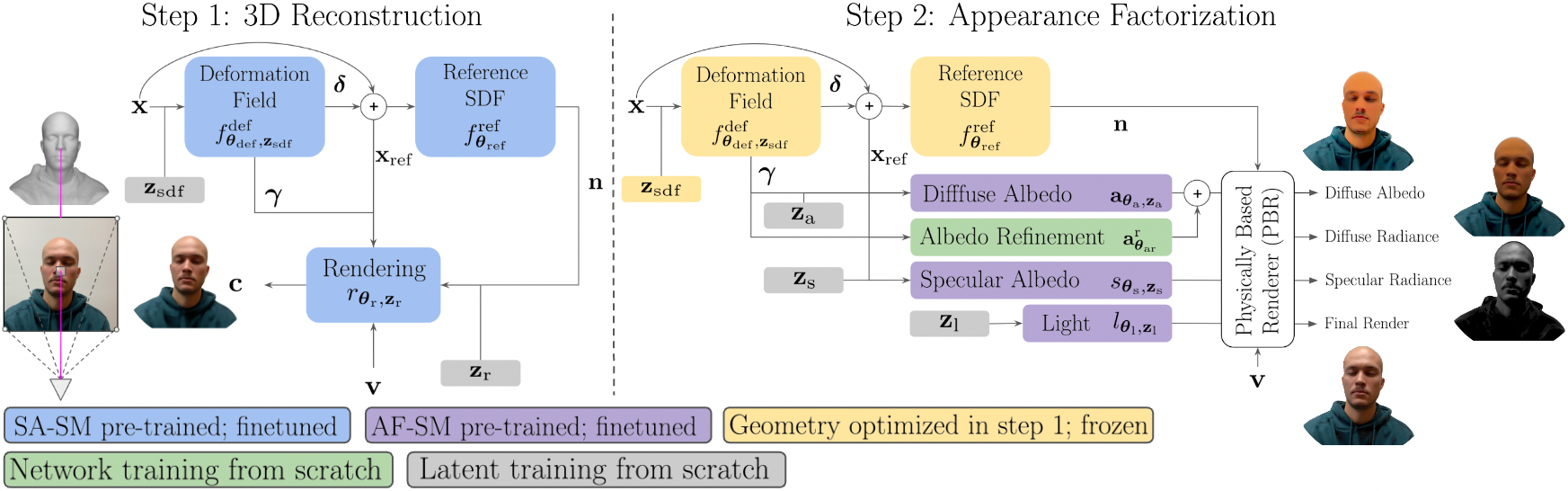}
    \caption{Inverse rendering using \method{}.}
    \label{fig:method}
    \vspace{-3mm}
\end{figure*}

\noindent\textbf{Neural fields for 3D reconstruction and scene factorization. } Recently, neural fields have been proposed as scene representations \cite{xie2021neural}, obtaining impressive results on  novel view synthesis \cite{mildenhall2020nerf,martin2021nerf,park2021nerfies,pumarola2021d,sitzmann2019scene} and 3D reconstruction \cite{niemeyer2020differentiable,ramon2021h3d,yariv2020multiview,zhang2021physg}, with application to modelling full head avatars \cite{park2021nerfies,park2021hypernerf,zheng2021avatar,gafni2021dynamic}. By combining surface priors \cite{park2019deepsdf} and surface rendering \cite{yariv2020multiview}, neural fields enable very accurate 3D reconstructions of the full head, including hair and shoulders \cite{ramon2021h3d,zheng2021avatar}. However, these methods require multi-view information and they do not recover the diffuse and specular albedos, and the illumination. Recently, novel priors that jointly model surface and appearance have been proposed \cite{yenamandra2021i3dmm}, but they have not been applied to 3D reconstruction from images. Neural fields have also been proposed for de-rendering scenes from several multi-view posed images \cite{zhang2021nerfactor,boss2021nerd,srinivasan2021nerv,boss2021neural,kuang2022neroic,sun2021nelf,zhang2021physg}. These methods combine neural fields with physically-based renderers, disentangling the intrinsic properties of the scene through direct supervision on  images. They also introduce priors on the materials to disambiguate the appearance factorization process \cite{zhang2021nerfactor}. However, it still remains a challenge to solve the inverse rendering problem from a single image using neural fields.

%% file: sections/3_method.tex
\vspace{-1mm}
\section{Method}

We follow an analysis-by-synthesis approach to retrieve all the components of a relightable avatar from a single posed image $\bI$ with associated foreground mask $\bM$, and camera parameters $\bC$. The  3D geometry is represented as a signed distance function (SDF) $\fsdf: \bx \rightarrow s$, such that the surface $\mathcal{S}$ is implicitly defined as $\mS = \{\bx \in \mbR^3 | \fsdf(\bx)=0\}$. To capture the complex appearance of human faces, we factor the surface radiance into a global illumination and spatially-varying diffuse and specular albedos, which we also implement as neural fields. All our neural fields are implemented as multilayer perceptrons. Find implementation details in the supplementary material.

Instead of using a single architecture to simultaneously optimize the geometry and appearance from scratch, we split the problem into a 3D reconstruction and an appearance factorization part. This two-step approach, similar to \cite{zhang2021nerfactor}, allows us to adapt the training scheme to each of the two problems independently, and to introduce appropriate inductive biases. These are key to resolve the ambiguities that exist in single-view 3D reconstruction and appearance factorization (Fig. \ref{fig:method}).

In a first step, we recover the 3D geometry of the scene from a single image.  \cite{yariv2020multiview,ramon2021h3d} have shown that $f^{\rm sdf}$ can be reconstructed from a collection of multiview posed images by using a differentiable rendering function $r: (\bx, \bn, \bv) \rightarrow \bc$ that models the radiance $\bc$ emitted from a surface point $\bx$ with normal $\bn$ in a viewing direction $\bv$, and minimizing a photometric error w.r.t. the input images. However, achieving similar reconstructions from a single view remains a challenge due to the lack of multi-view cues, which are important to disambiguate geometric and color information. We propose an architecture that yields accurate 3D reconstructions without requiring multi-view information by leveraging two main inductive biases: first, we decompose $\fsdf$ into a reference SDF and a deformation field \cite{yenamandra2021i3dmm}. We use this parameterisation as an implicit bias to constrain the composed SDF to be close to the reference. Second, we pre-train $\fsdf$ and $r$ to represent a shape and appearance statistical model (\priorgeo{}). Inspired by \cite{ramon2021h3d}, at inference time we optimise the parameters of this statistical model to obtain a robust initialisation for the analysis-by-synthesis process. These inductive biases greatly improve the performance of \method{} over \cite{ramon2021h3d} for the single view setup.

\vspace{-1mm}
In a second step, we factor the appearance of the reconstructed surface into a global illumination and spatially-varying diffuse and specular albedos using the physically-based renderer of \cite{zhang2021physg}.  To prevent shadows and specularities from being baked into the albedo, we first learn a statistical model of illumination and diffuse and specular albedos, or appearance factorization statistical model (\priorcolor{}), which we later use to constrain the search of appearance parameters to a suitable subspace. At inference time, we fit the parameters of this statistical model to a new scene, obtaining a coarse initialisation. To recover personalised details outside of the statistical model, we fine-tune the neural fields that encode the appearance. We find that at this point it is important to use regularisation losses and scheduling to prevent the illumination from leaking into the albedo.

\vspace{-2mm}
\subsection{Statistical models} 
We use a collection of scenes, composed of raw head scans paired with multiview posed images, to learn the statistical models for shape and appearance (\priorgeo{}), and appearance factorization (\priorcolor{}). For every scene, indexed by $i=1\dots M$, we have a set of surface points $\bx \in \mP_{\rm s}^{(i)}$ with associated normal vector $\bn$. We project each surface point to the images where it is visible, obtaining a set $\mC^{(i)}_\bx = \{(\bc, \bv)\}$ composed of pairs of associated RGB color $\bc$, and viewing direction $\bv$.

\vspace{-2mm}
\subsubsection{\priorgeo{} architecture}
We build a statistical model of shape and appearance, designed to enable the downstream task of single-view 3D reconstruction (Fig. \ref{fig:method}-left). Our architecture is composed of two main neural field decoders: an SDF decoder, $f^{\rm sdf}_{\btheta_{\rm sdf}, \bz_{\rm sdf}}$, and a non-physically-based rendering function decoder,  $r_{\btheta_{\rm r}, \bz_{\rm r}}$. Here, $\bz_{\rm sa} = \{\bz_{\rm sdf}, \bz_{\rm r}\}$ are the latent vectors of the shape and appearance spaces of the \priorgeo{}, and  $\btheta_{\rm sa} = \{\btheta_{\rm sdf}, \btheta_{\rm r}\}$ are the parameters of their respective decoders. 

\begin{figure}[t]
    \centering
    \includegraphics[width=1.0\columnwidth]{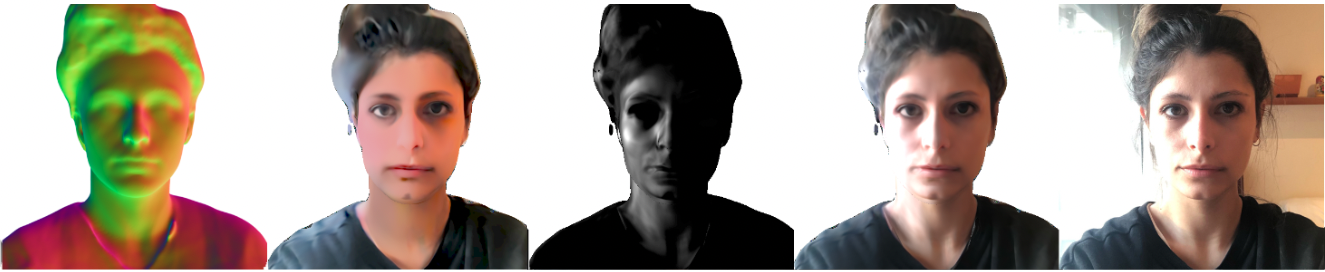}
    \caption{Inverse rendering of a scene under extreme illumination conditions using \method. Predictions from left to right: surface, normals, albedo, diffuse albedo, specular albedo and final rendering. Right-most: Input image.}
    \label{fig:irene}
    \vspace{-3mm}
\end{figure}

The SDF decoder is structured in two sub-functions: a deformation function and a reference SDF. As we show in the results, this separation acts as an implicit bias against staying too far from the reference SDF, which stabilizes the single-view 3D reconstructions. The deformation function, 

\vspace{-1mm}
\begin{equation}
    \label{eq:one_shot_def_net}
    f^{\rm def}_{\btheta_{\rm def}, \bz_{\rm sdf}}: \mathbb{R}^3 \rightarrow \mathbb{R}^{3+N_\gamma} \hspace{1mm},\hspace{1mm} \bx \mapsto (\bdelta, \bgamma),
\end{equation}
parameterised by internal parameters $\btheta_{\rm def}$ together with the latent vector $\bz_{\rm sdf}$, maps input coordinates, $\bx$, to a deformation 3-vector, $\bdelta$. It also outputs an auxiliary feature vector $\bgamma$ of dimension $N_\gamma$, which encodes higher level information required by the differentiable renderer \cite{yariv2020multiview}. The predicted deformation is used to map an input coordinate $\bx$ to a coordinate $\bx_{\rm ref}$ in a reference space where we evaluate a reference SDF $f^{\rm ref}_{\btheta_{\rm ref}}$ parameterized by internal parameters $\btheta_{\rm ref}$:

\vspace{-6mm}
\begin{subequations}
    \begin{align}
    \label{one_shot_canonical_coord}
    \bx_{\rm ref} &= \bx + \bdelta, \\
    f^{\rm ref}_{\btheta_{\rm ref}}&:\mathbb{R}^3\rightarrow \mathbb{R} \hspace{1mm} , \hspace{1mm} \bx_{\rm ref} \mapsto s.
    \end{align}
\end{subequations}

 Putting them together we obtain the composed SDF decoder
\begin{equation}
    \label{eq:one_shot_geo_net}
    f^{\rm sdf}_{\btheta_{\rm sdf}, \bz_{\rm sdf}} : \bx \mapsto f^{\rm ref}_{\btheta_{\rm ref}}(\bx^{\rm ref}),
\end{equation} 

where the decoder internal parameters are $\btheta_{\rm sdf} = (\btheta_{\rm def}, \btheta_{\rm ref})$. The second main component of our architecture is the rendering function, 
\begin{equation}
    \label{eq:one_shot_render_net}
    r_{\btheta_{\rm r}, \bz_{\rm r}}: (\bx_{\rm ref}, \bn, \bv, \bgamma) \mapsto \bc 
\end{equation}
parameterised by internal parameters $\btheta_{\rm r}$ and a latent vector $\bz_{\rm r}$. This function assigns an RGB color $\bc$, to every combination of 3D coordinate in the reference space $\bx_{\rm ref}$, unit normal vector $\bn$, and unit viewing direction vector $\bv$.

\vspace{-1mm}
\subsubsection{\priorgeo{} training}
 To train our \priorgeo{} we follow an auto-decoder framework, where each scene is assigned a set of latents $\bz_{\rm sa}^{(i)} = \{\bz_{\rm sdf}^{(i)}, \bz_{\rm r}^{(i)}\}$, which are optimized together with the statistical model parameters $\btheta_{\rm sa}$. After training, we obtain the parameters $\btheta_{\rm sa,0} = \{\btheta_{\rm sdf,0}, \btheta_{\rm r,0}\}$ such that any combination of latents $(\bz_{\rm sdf}, \bz_{\rm r})$ inside the latent space correspond to a well-behaved SDF, $f^{\rm sdf}_{\btheta_{\rm sdf,0}, \bz_{\rm sdf}}$, and appearance, $f^{\rm rend}_{\btheta_{\rm r,0}, \bz_{\rm r}}$, of a human head. We drop the dependence on the decoder internal parameters. 

To learn a space of head shapes, for each scene, we sample a set of points on the surface, $\mP_{\rm s}^{(i)}$, and compute the surface error loss $\mathcal{L}_{\rm Surf}^{(i)} = \sum_{\bx_j \in \mathcal{P}_{\rm s}^{(i)}} |f^{\rm sdf}_{\bz^{(i)}_{\rm sdf}}(\bx_j)|$. We also sample another set uniformly taken from the scene volume, $\mP_{\rm v}^{(i)}$, and compute the Eikonal loss \cite{gropp2020implicit}$~
    \mathcal{L}_{\rm Eik}^{(i)} = \sum_{\bx_k \in \mathcal{P}_{\rm v}^{(i)} }  (\lVert\nabla_{\bx} f^{\rm sdf}_{\bz^{(i)}_{\rm sdf}}(\bx_k)\rVert-1 )^2.$
We promote small-magnitude and zero-mean deformations, which avoids solutions where the deformations compensate for an unnecessarily offset or scaled reference SDF:
\vspace{-1mm}
\begin{equation}
\begin{split}
\mL_{\rm Def}^{(i)} = \frac{1}{|\mP_{\rm s}^{(i)}|} \biggl( \sum_{\bx_j \in \mP_{\rm s}^{(i)}} \lVert \bdelta_{j}^{(i)}\rVert_2 + \biggl\lVert{\sum_{\bx_j \in \mP_{\rm s}^{(i)}} \bdelta_{ j}^{(i)}}\biggr\rVert_2\biggr),
\end{split}
\end{equation}
where $\bdelta_{j}^{(i)}$ is the deformation vector applied to the 3D point $\bx_j$ of the scene $i$.  

Similarly to \cite{yenamandra2021i3dmm}, we use a landmark consistency loss. We automatically annotate a set of 3D face landmarks $\{\bx_l^{(i)}\}$ with $l=1 \dots L$ for each scene, $i$, and use their deformed coordinate mismatch between pairs of scenes as a loss, $\mL_{\rm Lm}^{(i)} = \sum_{j\neq i} \sum_{l}^{L} \lVert\bx_{\rm ref,l}^{(i)} - \bx_{\rm ref,l}^{(j)}\rVert^2$, 
where $\bx_{\rm ref,l}^{(i)}$ is the position of landmark $l$ of scene $i$ in the reference space.

The \priorgeo{} learns a distribution of head appearances from the set of posed images accompanying every training scene. To evaluate the rendering function (eq. \ref{eq:one_shot_render_net}), we compute the coordinate in the reference space $\bx_{\rm ref}$ associated to the surface point (eq. \ref{one_shot_canonical_coord}), as well as the high-level descriptor $\bgamma$ (eq. \ref{eq:one_shot_def_net}). We also obtain the surface normals, $\bn$, as the normalized gradient of the SDF \cite{yariv2020multiview}. With these, we define the color loss
\begin{equation}
    \label{eq:geometry_prior_color_loss}
    \mL_{\rm Col}^{(i)} = \sum_{\bx \in \mP_{\rm s}^{(i)}}\sum_{(\bc, \bv)\in\mC_\bx^{(i)}}  \lVert r_{\bz^{(i)}_{\rm r}}(\bx_{\rm ref}, \bn, \bv, \bgamma) - \bc \rVert .
\end{equation}
Finally, $\mathcal{L}_{\rm Emb}^{(i)}$ enforces a zero-mean multivariate-Gaussian distribution
with spherical covariance $\sigma^2$ over the spaces of shape and appearance latent  vectors: 
$\mathcal{L}_{\rm Emb}^{(i)} = \frac{1}{\sigma^{2}}\bigl(\lVert\bz^{(i)}_{\rm sdf} \rVert_2 + \lVert\bz^{(i)}_{\rm r} \rVert_2\bigr).$
Putting them together, we minimize the following:

\vspace{-3mm}
\begin{equation} \label{eq:deep_sdf_objective_color}
\begin{split}
\argmin_{\{\bz^{(i)}_{\rm sa}\},  \btheta_{\rm sa}} \sum_i \mL_{\rm Surf}^{(i)}  + \lambda_1\mL_{\rm Eik}^{(i)} + \lambda_2\mL_{\rm Def}^{(i)}+ \lambda_3\mL_{\rm Lm}^{(i)} + \\ \lambda_4\mL_{\rm Col}^{(i)}  +  \lambda_5\mL_{\rm Emb}^{(i)}
\end{split}
\end{equation}
\vspace{-1mm}
where $\lambda_{1-5}$ are scalar hyperparameters.

\vspace{-1mm}
\subsubsection{\priorcolor{} architecture}
We build a statistical model of illumination and materials to enable the downstream task of single-view appearance factorization (Fig. \ref{fig:method}-right). We use the physically-based differentiable rendering model introduced in \cite{zhang2021physg}  
to capture the complex appearance of human faces, which can include shadows and view-dependent specular reflections \cite{weyrich2006analysis}.

We compute the radiance $r^{\rm pb}$ emitted from a surface point $\bx$ with normal $\bn$ in the viewing direction $\bomega_{\rm o}$ using the non-emitting rendering equation 
\vspace{-1mm}
\begin{equation}
\begin{split}
    \label{eq:rendering_equation}
    & r^{\rm pb}(\bomega_{\rm o}, \bx)  = r^{\rm d}(\bomega_{\rm o}, \bx) +  k_{\rm s} ~ r^{\rm s}(\bomega_{\rm o}, \bx) = \\ & = \int_{\Omega} l(\bomega_{\rm i}) \big(f^{\rm d}(\bx) + k_{\rm s} ~ f^{\rm s}(\bx, \bomega_{\rm i}, \bomega_{\rm o})\big) (\bomega_{\rm i} \cdot \bn ) {\rm d}\bomega_{\rm i},
\end{split}
\end{equation}
 
where $l(\bomega_{\rm i})$ is the incident light from direction $\bomega_{\rm i}$, the functions $f^{\rm d}, f^{\rm s}$ are the diffuse and specular components of the BRDF respectively, and the scalar $k_{\rm s}\in [0,1]$ controls their relative weight. The funcions $r^{\rm d}$ and $r^{\rm s}$ represent the integrated radiance corresponding to the diffuse and specular parts of the BRDF respectively. The integral is computed over the hemisphere $\Omega = \{\bomega_{\rm i}: \bomega_{\rm i}\cdot \bn > 0\}$. 

\begin{figure}[t]
    \centering
    \includegraphics[width=\columnwidth]{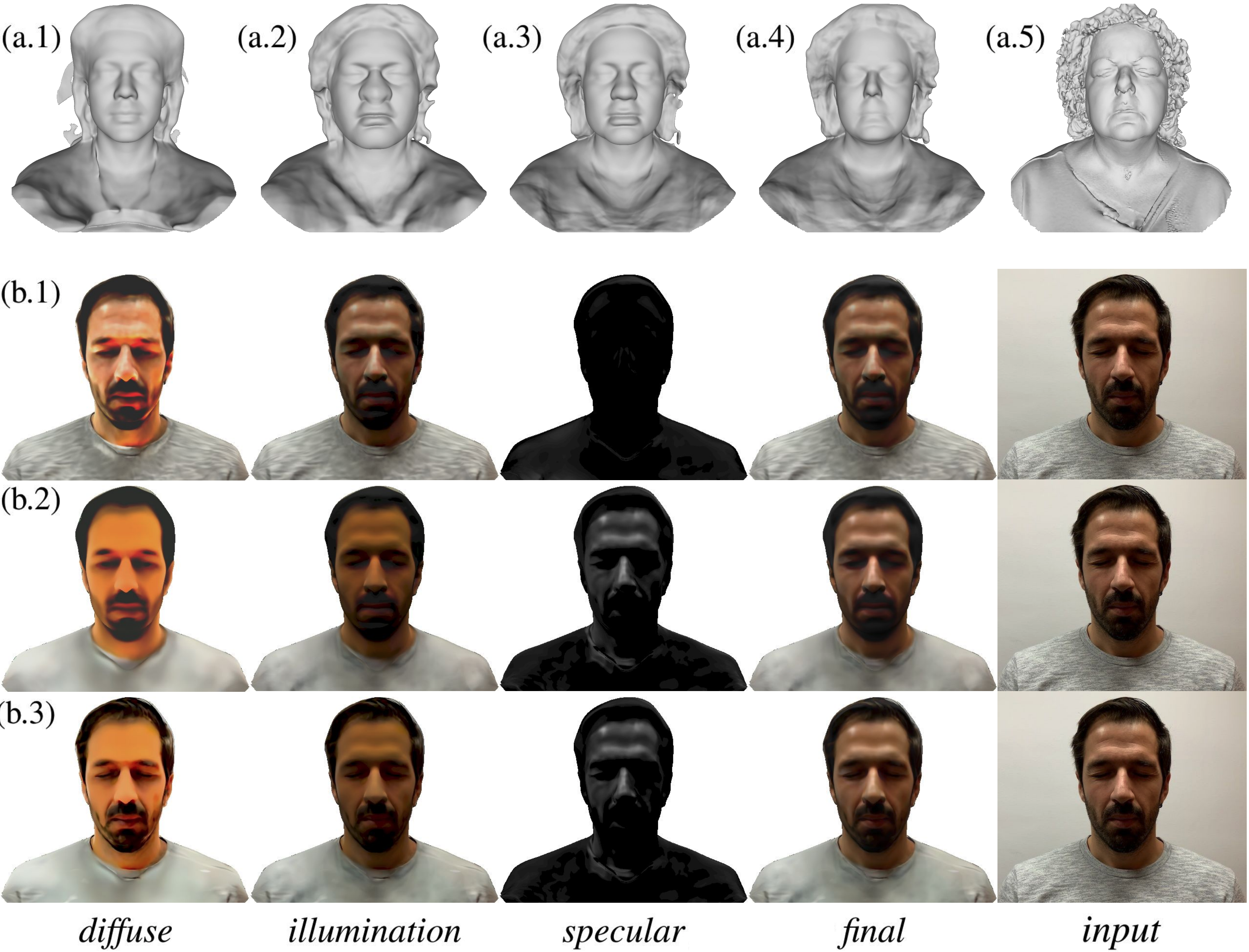}
    \caption{
    {\bfseries Ablation study:} In (a) we ablate the 3D reconstruction method. In (b) we ablate the inverse-rendering method. See text for details.}

    \label{fig:ablation_color}
    \vspace{-3mm}
\end{figure}

Following \cite{zhang2021physg} and \cite{wang2009all}, we approximate the incident light, specular reflectance, and clamped cosine decay factor, with spherical gaussian  decompositions, which allows us to efficiently compute the integral in  closed form. We represent the environment map $l(\bomega_{\rm i})$ as a mixture of $N_{\rm l}$ spherical gaussians. The diffuse component of the BRDF is a scaled spatially varying RGB albedo, $\ba\in\mbR^3$, with no angular dependence:  $f^{\rm d}(\bx) = \ba(\bx)/\pi$. As for the specular component of the BRDF, $f^{\rm s}$, we particularise the simplified Disney BRDF \cite{karis2013real} used in \cite{zhang2021physg} by fixing the value of the roughness parameter. For a given point in the surface and a given viewing direction, having fixed the roughness, the only free parameter of $f^{\rm d}$ is a spatially-varying monochrome specular albedo $s(\bx) \in \mbR$. We provide more details about the rendering model in the supplementary material.

Determining the appearance of a scene then boils down to regressing the spatially-varying diffuse and specular albedos, $\ba$ and $s$, as well as the lighting parameters $\{\bxi_l, \lambda_l, \bmu_l\}$, where $\bxi_l\in\mbS^2$ is the direction of the lobe, $\lambda_l\in\mbR_+$ is the lobe sharpness, and $\bmu_l\in\mbR^{n}_{+}$ the lobe amplitude. We represent the \priorcolor{} with three decorders. First, a diffuse and specular albedo decoders, 

\vspace{-3mm}
\begin{subequations}
\label{eq:albedo_decoders}
\begin{align}
\ba(\bx) &= \ba_{\btheta_{\rm a}, \bz_{\rm a}}(\bx_{\rm ref}, \bgamma): \mbR^3 + N_\gamma \rightarrow \mbR^3 \\ s(\bx) &= s_{\btheta_{\rm s}, \bz_{\rm s}}(\bx_{\rm ref}): \mbR^3 \rightarrow \mbR,
\end{align}
\end{subequations}
parameterised respectively by the internal parameters $\btheta_{\rm a}$ and $\btheta_{\rm s}$, and with associated latent vectors $\bz_{\rm a}$ and $\bz_{\rm s}$. These latent vectors are decoded to a space of continuous albedo functions, which we evaluate and supervise at the surface points. Instead of using the raw surface points as inputs to the decoders, we first process them with $f^{\rm def}$, initialised with the parameters $\btheta_{\rm def, 0}$ and latent vectors corresponding to the geometry of each scene in the collection, $\bz_{\rm sdf}^{(i)}$, to obtain their correspondences in the learnt reference space, $\bx_{\rm ref}$ (eq. \ref{one_shot_canonical_coord}), as well as their associated descriptor $\bgamma$ (eq. \ref{eq:one_shot_def_net}). We provide these to the \priorcolor{} to allow it to re-use the semantic information $\bgamma$ extracted by $f^{\rm def}$ for each surface point.  In addition, we have an illumination decoder $l_{\btheta_{\rm l},\bz_{\rm l}}\in\mbR^{N_{\rm l} \times 7}$, parameterised by $\btheta_{\rm l}$ and with associated latent vector $\bz_{\rm l}$, which outputs the parameters of the $N_{\rm l}$ spherical gaussians that describe the environment map of a scene.  

The color associated to surface point $\bx$ seen from viewing direction $\bomega_{\rm o}$ is computed by evaluating the physically-based rendering equation (eq. \ref{eq:rendering_equation}) with the parameters obtained from these decoders. Here, we denote this rendered color by $r^{\rm pb}_{\btheta_{\rm pb}, \bz_{\rm pb}}(\bx, \bomega_{\rm o})$, where $\btheta_{\rm pb} = \{\btheta_{\rm a}, \btheta_{\rm s}, \btheta_{\rm l}\}$, and $\bz_{\rm pb} = \{\bz_{\rm a}, \bz_{\rm s}, \bz_{\rm l}\}$.

\begin{figure}[t]
    \centering
    \includegraphics[width=\columnwidth]{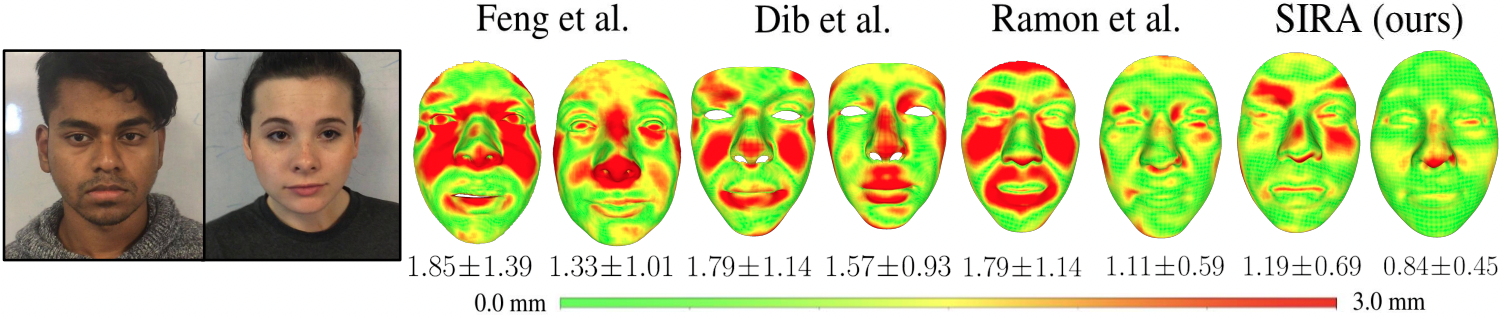}
    \caption{{\bfseries Single-view 3D reconstruction:} Subjects from 3DFAW dataset \cite{pillai20192nd}. Comparison: Feng 2021 \cite{feng2021learning}, Dib 2021 \cite{dib2021towards}, Ramon 2021 \cite{ramon2021h3d}, \method{} (ours).}
    \label{fig:quantitative}
    \vspace{-3mm}
\end{figure}

\vspace{-2mm}
\subsubsection{\priorcolor{} training}

These decoders are trained in an auto-decoder setup, assigning latent vectors $\bz_{\rm pb}^{(i)}$ to each scene.
The appearance factorization model is trained in a self-supervised manner, since only multi-view images with known cameras are assumed.
After training, we obtain the optimized parameters $\btheta_{\rm pb,0}= \{\btheta_{\rm a,0}, \btheta_{\rm s,0}, \btheta_{\rm l,0}\}$ such that any combination of latents $\bz_{\rm pb}$ inside the latent space correspond to a well-behaved illumination, as well as diffuse and specular albedos. During training, we set the parameter that controls the relative weight of the diffuse and specular components to $k_{\rm s} = 1$, and minimize the following loss:

\begin{figure}[t!]
    \centering
    \includegraphics[width=\columnwidth]{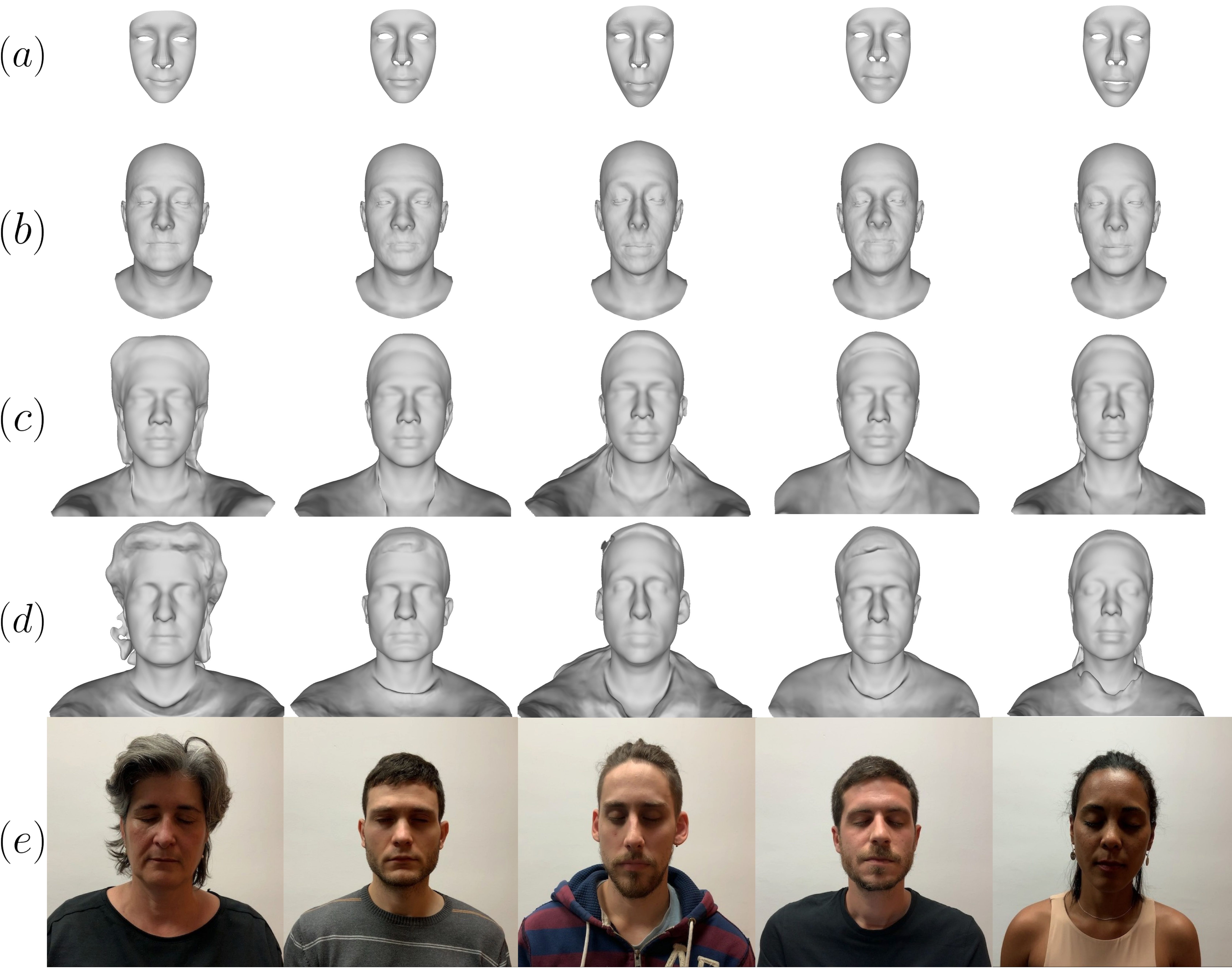}
    \vspace{-6mm}
    \caption{{\bfseries Single-view 3D reconstruction:} Subjects from the H3DS dataset. (a) Dib 2021 \cite{dib2021towards}, (b) Feng 2021 \cite{feng2021learning}, (c) Ramon 2021 \cite{ramon2021h3d}, (d) \method (Ours) and (e) input image.}
    \label{fig:qualitative_full_head}
    \vspace{-3mm}
\end{figure}
\vspace{-3mm}

 \begin{figure*}[t]
    \centering
    \includegraphics[width=\textwidth]{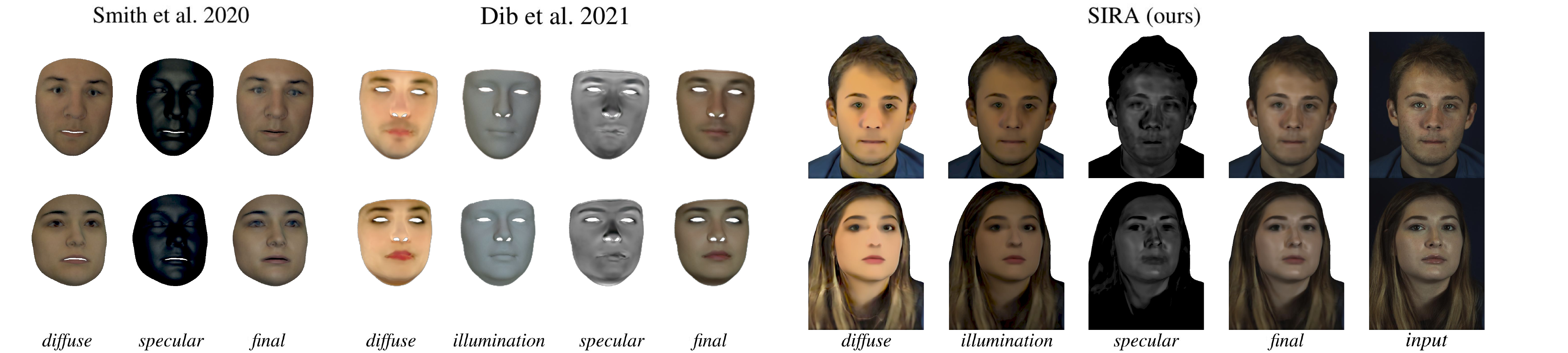}
    \vspace{-6mm}
    \caption{\textbf{Inverse rendering:} Subjects from 3DFAW-HR dataset. Comparison: Smith 2020 \cite{smith2020morphable}, Dib 2021 \cite{dib2021towards}, and \method (ours). \textit{Input} images are decomposed into \textit{diffuse} albedo, diffuse \textit{illumination}, \textit{specular} radiance, and \textit{final} render} 
    \label{fig:disentanglement}
    \vspace{-5mm}
\end{figure*}

\begin{equation}
    \argmin_{\{\bz_{\rm pb}^{(i)}\}, \btheta_{\rm pb}} \mL_{\rm Col}^{(i)} + \lambda_{6} \mL_{\rm Emb}^{(i)} +  \lambda_{7} \mL_{\rm Reg}^{(i)} 
\end{equation}
where $\lambda_6$ and $\lambda_7$ are hyperparameters 
The first component of the loss, $\mL_{\rm Col}^{(i)}$, is the photometric error between the color rendered by $r^{\rm pb}$ and the ground truth images from different views. It is defined analogously to eq. \ref{eq:geometry_prior_color_loss}. We also use a latent vector regularization, $\mL_{\rm Emb}$, defined analogously to the equivalent loss used to train the \priorgeo{}. Finally, to avoid baking shadows and reflections in the diffuse albedo, we include a regularisation loss that encourages it to be spatially smooth: $\mL_{\rm Reg}^{(i)} = \sum_{\bx\in \mP_{\rm s}^{(i)}} \lVert \ba(\bx) - \ba(\bx + \bepsilon)\rVert$
where $\bepsilon$ is a 3D perturbation set as a hyperparameter. 

 \vspace{-1mm}
\subsection{Single-view inverse rendering}

With our pre-trained statistical models at hand, we can tackle the task of obtaining a 3D reconstruction and a factorized appearance from a single portrait image $\bI$ with associated camera parameters $\bC$ and foreground mask $\bM$. 

\vspace{-1mm}
\subsubsection{Reconstructing geometry from a single image}
To obtain 3D reconstructions of new scenes, we render the geometry described by $f^{\rm sdf}$ using the differentiable rendering function $r$ of eq. \ref{eq:one_shot_render_net}, and minimize a photoconsistency error. For a pixel coordinate $p$ of the input image $\bI$, we march a ray $\br = \{\bc + t\bv | t \geq 0\}$,  where $\bc$ is the position of the associated camera $\bC$, and $\bv$ the viewing direction. We find the intersection coordinates with the composed SDF (eq. \ref{eq:one_shot_geo_net}) using sphere tracing. This intersection point can be made differentiable w.r.t $\bz_{\rm sdf}$ and $\btheta_{\rm sdf}$ using implicit differentiation \cite{niemeyer2020differentiable,yariv2020multiview}. The differentiable intersection coordinates $\bx_{\rm s}$ are used to obtain their associated 3D displacement $\bdelta$ and  feature vector $\bgamma$ (eq. \ref{eq:one_shot_def_net}), as well as their corresponding coordinates in the reference space $\bx_{\rm ref}$ (eq. \ref{one_shot_canonical_coord}), and normal vector $\bn = \nabla_\bx f^{\rm sdf}$. Then, the color associated to the ray is computed as $\bc = r(\bx_{\rm ref}, \bn, \bv, \bgamma)$.

In order to optimize $\bz_{\rm sa}$ and $\btheta_{\rm sa}$, we minimize the same photoconsistency, mask, and eikonal losses as in \cite{yariv2020multiview}. Se supplementary material for a more detailed explanation.

Instead of optimizing all the parameters $\{\btheta_{\rm sdf}, \btheta_{\rm r}, \bz_{\rm sdf}, \bz_{\rm r}\}$ at once, we propose a two-step schedule, better suited for the underconstrained one-shot scenario. We initialise the geometry and rendering functions with the parameters obtained with the pretraining described in the last section, $\{\btheta_{\rm sdf,0}, \btheta_{\rm r,0}\}$. The initial shape and appearance latents, $\bz_{\rm sdf}$ and $\bz_{\rm r}$, are picked from a multivariate normal distribution with zero mean and small variance, ensuring that they start near the mean of the latent spaces. In a first optimization phase, we only optimize the shape and appearance latents. This yields an initial approximation within the previously learnt shape and appearance latent spaces. In a second phase, we unfreeze the parameters of the deformation and rendering nets, $\{\btheta_{\rm def}, \btheta_{\rm r}\}$ (eqs. \ref{eq:one_shot_def_net}, \ref{eq:one_shot_render_net}), but not those of the reference SDF, $\btheta_{\rm ref}$.

This schedule is key to obtaining accurate results in the one-shot regime. While unfreezing the deformation and rendering networks allows us  to reach highly-detailed solutions outside of the pre-learnt latent spaces, the fact that we express the shape as a deformed reference SDF acts as a regularization that allows correct training convergence. We refer to the fine-tuned shape parameters as $\btheta_{\rm def,ft}$ and $\bz_{\rm sdf, ft}$.

\vspace{-2mm}
\subsubsection{Appearance factorization from a single image}

Once $f^{\rm sdf}$ has been optimized for image $\bI$, we dispose of the non-disentangled renderer $r_{\btheta_{\rm r}, \bz_{\rm r}}$, and tackle the appearance factorization problem. We use ray marching to obtain the 3D surface coordinates $\bx$, corresponding to each pixel $p\in\mP$ with a non-zero foreground mask value. We process these coordinates with $f^{\rm def}$ to compute their correspondences in the reference space, $\bx_{\rm ref}$, and associated descriptors $\bgamma$, which are the inputs to the \priorcolor{}.

To better capture personalized details for each scene, outside of the pretrained latent space, 
we extend the \priorcolor{} with a diffuse albedo refinement module. We express the diffuse component of the BRDF (eq. \ref{eq:albedo_decoders}) as:

\vspace{-2mm}
\begin{equation}
    \ba(\bx) = \ba_{\btheta_{\rm a}, \bz_{\rm a}}(\bx_{\rm ref}, \bgamma) + k_{\rm r} ~ \ba^{\rm r}_{\btheta_{\rm ar}}(\bx_{\rm ref}, \bgamma)
    \label{eq:albedo_refine}
    \vspace{-1mm}
\end{equation}
where $\ba^{\rm r}$ is an albedo refinement neural field parameterized by $\btheta_{\rm ar}$. The scalar $k_{\rm r}$ controls the weight of the albedo refinement field. This separation of the albedo into a base and refinement fields enables adding detail to the coarse albedo provided by the pretrained statistical model $\ba_{\btheta_{\rm a}, \bz_{\rm a}}$, while using regularization losses that prevent the refinement $\ba^{\rm r}$ from absorbing shading and specular information. In this section, we denote the rendered color using this modified model as $r^{\rm pb}(\bx, \bomega_{\rm o})$, dropping the dependence on its internal parameters $\btheta_{\rm pb} = \{\btheta_{\rm a}, \btheta_{\rm ar}, \btheta_{\rm s}, \btheta_{\rm l}\}$ and latent vectors $\bz_{\rm pb} = \{\bz_{\rm a}, \bz_{\rm s}, \bz_{\rm l}\}$.

To optimize $r^{\rm pb}$, we minimize the loss $\mL = \mL_{\rm RGB} + \mL_{\rm Reg}$. The rendering photoconsistency loss $\mathcal{L}_{\rm RGB} = |\mP|^{-1} \sum_{p\in \mathcal{P}} |\bI(p) - \bc(p)|$ is defined on physically-based rendered color $\bc = r^{\rm pb}(\bx, \bomega_{\rm o})$ evaluated at the coordinates $\bx$ and viewing directions $\bomega_{\rm o}$ corresponding to pixel $p$. We introduce a regularization loss, $\mL_{\rm Reg}$, designed to prevent the albedo refinement from explaining color variations that should be captured by the diffuse or specular shading. This loss is defined as 

\vspace{-4mm}
\begin{equation}
    \label{eq:albedo_refinement_regularization}
    \mL_{\rm Reg} = \frac{1}{|\mP|}\sum_{p\in\mP} \lVert ~ \ba^{\rm r}(p)\lVert w(p)
\end{equation}

\vspace{-4mm}
\begin{equation}
     w =\lambda_{8} \max \big(0, \lVert \ba_{\btheta_{\rm a}, \bz_{\rm a}} \rVert_1 - \lVert r^{\rm d}_{\rm b}\rVert_1 \big)+ \lambda_{9} \lVert r^{\rm s}\rVert_1
\end{equation}
where $r^{\rm s}$ is the specular component of the radiance in the rendering equation (eq.  \ref{eq:rendering_equation}), and $r^{\rm d}_{\rm b}$ is the diffuse component of the radiance evaluated with the base albedo $\ba_{\btheta_{\rm a}, \bz_{\rm a}}$. The scalars $\lambda_{8}$ and $\lambda_{9}$ are hyperparameters.  The weighting function $w$ has been designed to have high values for pixels where there are shadows or reflections. By regularizing the norm of $\ba^{\rm r}$ for those pixels (eq. \ref{eq:albedo_refinement_regularization}), it prevents the albedo refinement field from absorbing shading and specular information.

We do not optimize all the parameters at once, as this would tend to bake shadows and specularities into the albedo. To prevent this from happening, we design a suitable scheduling of the learning rates, as well as of the rendering model hyperparameters $k_{\rm s}$ and $k_{\rm r}$. Roughly, our scheduling follows this order: we first learn illumination, recovering coarse shadows on a fixed initial albedo, without specularities or albedo refinement ($k_{\rm s}=k_{\rm r}=0$). Then, we optimize $\bz_{\rm a}$, allowing the model to learn an albedo within the latent space.  Next, we gradually add and optimize specular reflections ($k_{\rm s}=1)$. With this in place, we freeze the coarse albedo and unfreeze the albedo refinement module. During this stage, the model captures photo-realistic details in the albedo refinement field, while avoiding baking shades and reflections thanks to $\mL_{\rm Reg}$. A more fine-grained description of our training schedule can be found in the supplementary material.

%% file: sections/4_experiments.tex
\vspace{-1mm}
\section{Experiments}
We next evaluate our 3D reconstruction and appearance factorization 
on multiple real-world portrait photos from the datasets H3DS \cite{ramon2021h3d}, 3DFAW \cite{pillai20192nd} and Wikihuman Project \cite{WinNT}. We train the SA-SM and AF-SM priors on the dataset used in \cite{ramon2021h3d}. See supplementary material for a more detailed explanation of the training and evaluation datasets.
Training the priors and fitting \method{} for a scene takes about 1 day and 10 min respectively, on a single RTX 2080Ti GPU.

\vspace{-1mm}
\subsection{Ablation}

We conduct an ablation study on the H3DS dataset and show the qualitative results in Figure \ref{fig:ablation_color} for both 3D reconstruction and appearance factorization.

\noindent{\bf 3D reconstruction.} 
We select as baseline the architecture proposed in \cite{ramon2021h3d}. Note in Fig.~\ref{fig:ablation_color}-top, that this architecture underfits the scene when only the latent vector is optimized (a.1) and it is unstable when the decoder is fine-tuned (a.2). By splitting the geometry into a deformation field and a reference sdf (a.3), we gain more control over the resulting 3D surfaces, which leads to more plausible and stable solutions even when the deformation decoder is fine-tuned. Finally, by jointly modelling a distribution of 3D shapes and appearances with the \priorgeo{}, \method{} (a.4) is able to better disentangle geometric and visual information, providing 3D models that highly resemble to input image. The qualitative  results of Fig.~\ref{fig:ablation_color} are aligned with the errors reported in Table~\ref{quantitative_tables} (top), in which \method{} outperforms our ablated baselines by a significant margin.

\noindent{\bf Appearance factorization.} As shown in Fig.~\ref{fig:ablation_color}-right, directly fitting the physically-based rendering model of \method{} to a scene, without introducing any prior, yields baked lights and shadows in the diffuse albedo, and fails to recover a meaningful specular component (b.1). Using the \priorcolor{}, together with a scheduling to guide the optimization (b.2), we correctly factor the appearance components. However, the results lack  realism due to the albedo smoothness bias in the prior. To get sharper results (b.3), we introduce the albedo refinement module (Eq.~\ref{eq:albedo_refine}).

\begin{figure}[t]
    \centering
    \includegraphics[width=\columnwidth]{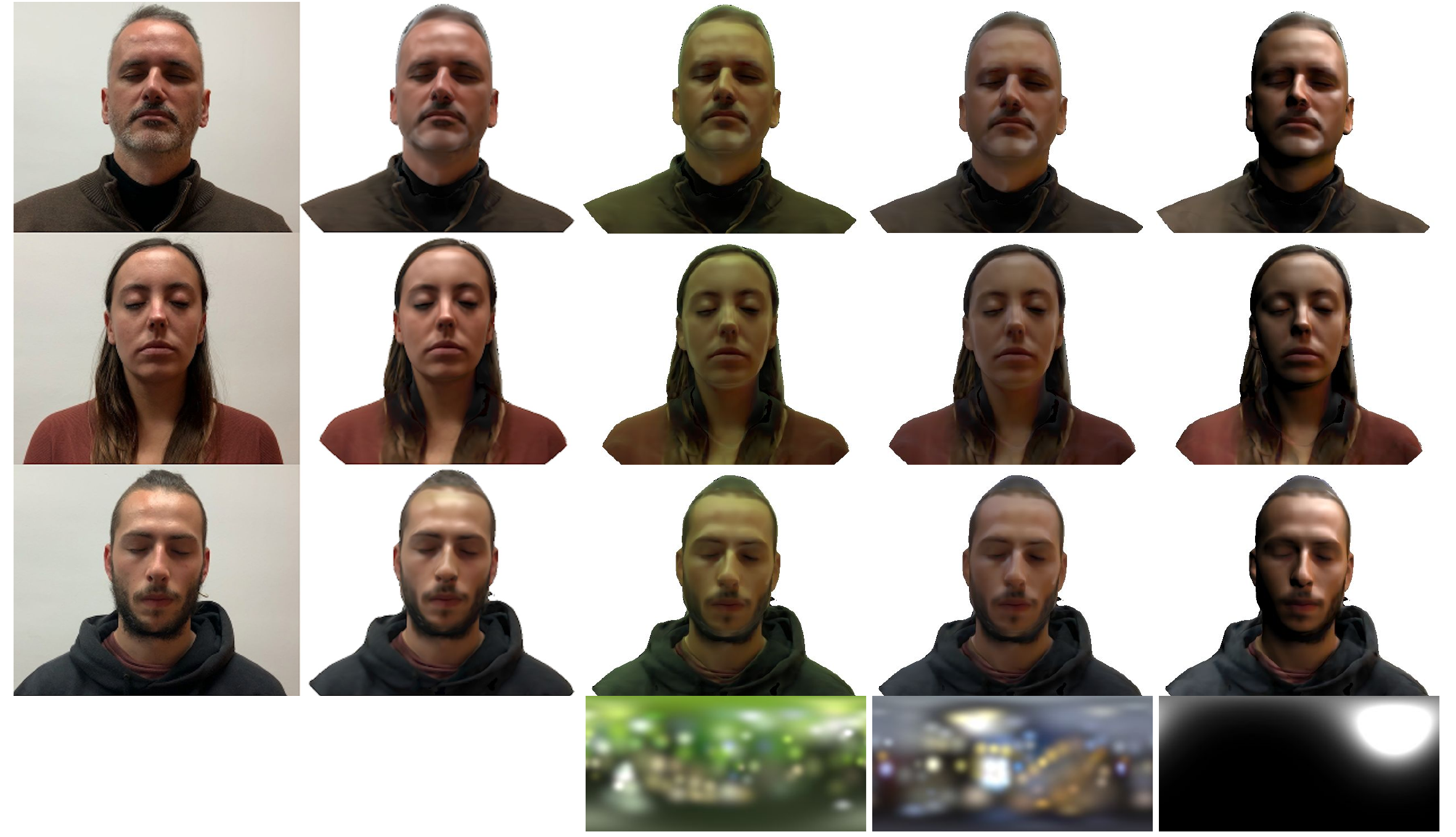}
    \vspace{-6mm}
    \caption{\textbf{Relighting} of inverse-rendered scenes. Subjects from the H3Ds dataset. 
    }
    \label{fig:relit_ours}
    \vspace{-3mm}
\end{figure}

\vspace{-1mm}
\subsection{3D reconstruction comparison}
We compare \method{} against the 3DMM-based methods Feng et al. 2021~\cite{feng2021learning} and Dib et al. 2021~\cite{dib2021towards}, as well as the unconstrained method Ramon et al. 2021~\cite{ramon2021h3d}.

Table \ref{quantitative_tables} (top) reports  the surface error in the facial area, using the unidirectional Chamfer distance from the predictions to the ground truth. \method{} achieves comparable results in all the datasets, outperforming the baselines on the 3DFAW low resolution subset and in H3DS. Moreover, as  shown in Fig.~\ref{fig:quantitative}, \method{} provides finer anatomical details in complex areas like the cheeks and  nose.

Unlike 3DMM-based approaches \cite{feng2021learning,dib2021towards}, \method{} recovers the geometry of the head, including hair and shoulders. This has an important perceptual impact, visible in Figure \ref{fig:qualitative_full_head}. While~\cite{ramon2021h3d}  reconstructs the same area, it underfits the input data. \method{}, in contrast, yields 3D shapes that clearly retain the identity of the person, yielding realistic relightable avatars.

\vspace{-1mm}
\definecolor{rowblue}{RGB}{220,230,240}
\begin{table}[t]

\centering

\setlength{\tabcolsep}{4pt} 
\vspace{2mm}
\centering
\rowcolors{3}{rowblue}{white}
\resizebox{0.75\columnwidth}{!}{

\begin{tabular}[t]{lccc}
\toprule
& 3DFAW $\downarrow$ & 3DFAW HR $\downarrow$ & H3DS $\downarrow$\\
\cmidrule{1-4}

Feng \cite{feng2021learning} & 1.53 & \bfseries 1.46  & 1.62 \\
Dib \cite{dib2021towards} & 1.53 & 1.61 & 1.70\\
Ramon \cite{ramon2021h3d} & 1.75 & 1.48 & 1.98\\
\cmidrule{1-4}
Ablation. Fig \ref{fig:ablation_color} a.2 & - & - & 2.17\\
Ablation. Fig \ref{fig:ablation_color} a.3 & - & - & 1.86\\
\method (Ours) & \bfseries 1.42 & 1.58 & \bfseries 1.46  \\
\bottomrule
\end{tabular}
}
\setlength{\tabcolsep}{2pt} 
\centering
\rowcolors{3}{rowblue}{white}
\resizebox{1.0\columnwidth}{!}{

\begin{tabular}[t]{lccccccc}
\toprule
& Final & Final & Diffuse & Diffuse & Specular & Specular \\
& (SSIM) $\uparrow$ & (PSNR) $\uparrow$& (SSIM) $\uparrow$& (PSNR) $\uparrow$& (SSIM) $\uparrow$& (PSNR) $\uparrow$\\
\cmidrule{1-7}
Lattas \cite{lattas2020avatarme} & - & - & 0.83 & 21.7 & 0.58 & 17.5\\
Yamaguchi\cite{yamaguchi2018high} & - & - & 0.85 & \bfseries 22.7 & \bfseries 0.71 & \bfseries 19.5\\
Dib \cite{dib2021towards} & 0.91 & 27.5 & 0.83 & 20.0 & 0.62 & 14.6 \\
Smith \cite{smith2020morphable} & 0.89 & 26.3 & 0.64 & 12.76 & 0.26 & 4.27\\
\method (Ours) & \bfseries 0.95 & \bfseries 33.4 & \bfseries 0.87 & 22.5 & 0.53 & 9.91\\
\bottomrule
\end{tabular}
}
\caption{{\bfseries (top) 3D reconstruction:} Average surface error in millimeters. \textbf{(bottom) De-rendering:} Evaluated on the Digital Emily scene. PSNR  in dB.
}
\label{quantitative_tables}
\vspace{-3mm}
\end{table}

\vspace{-1mm}
\subsection{Appearance factorization comparison}

We next analyze our results for the task of appearance factorization.
Fig.~\ref{fig:disentanglement}, shows qualitative comparison of \method{}  against Smith et al. 2020 \cite{smith2020morphable} and Dib et al. 2021 \cite{dib2021towards} on three cases from the 3DFAW high resolution dataset.  \method{} performs similar to the  baselines in the face region and additionally factorizes the intrinsic components of the hair and the upper body. Note that the appearance of the whole head is much more complex and diverse than the appearance of the skin in the facial region. Furthermore, \method{} models high frequency skin specularities, leading to photorealistic re-rendered images. This can be seen in Fig.~\ref{fig:relit_ours}, where three avatars reconstructed with \method{} are relighted under novel lighting conditions. Note how the specularities move in the forehead when the light moves around the head (columns 5-7). Finally, it is also worth mentioning that \method{} is robust to extreme illumination conditions of the input image. This is shown in Fig.~\ref{fig:irene}, where the input image is highly saturated by the scene illumination, but \method{} is still able to recover the intrinsic components correctly.

We report a quantitative analysis in Table \ref{quantitative_tables} (bottom), using the Digital Emily scene. Compared to the baselines \cite{dib2021towards,smith2020morphable,yamaguchi2018high,lattas2020avatarme}, we achieve slightly better results in the final render, and comparable results in the diffuse and specular albedos, while being the only method that recovers the appearance of the entire head.

%% file: sections/5_conclusions.tex
\vspace{-2mm}
\section{Conclusions}
We have introduced \method{}, the first approach for building 3D avatars of human heads from a single image which, besides reconstructing high fidelity geometry, allows factorizing surface materials and global scene illumination. In order to tackle such an under-constrained problem, we have introduced two novel statistical models based on neural fields that encode shape and appearance into low dimensional latent spaces. A thorough evaluation has shown that \method{} provides SOTA results on full head geometry reconstruction, while also disentangling global illumination, and diffuse/specular albedos, yielding 3D relightable avatars from one single portrait image. Next avenues in this topic include speeding up the optimization process of neural fields. 

\vspace{-2mm}
\paragraph{Ethical considerations:} Highly accurate photorealistic reconstructions can lead to identity impersonation concerns or image alteration. In addition, it is essential that the model is not biased and does not discriminate against any group, religion, colour, gender, age, or disability status. We include in the supplementary material the results on the Celeb-HQ dataset \cite{CelebAMask-HQ} to show that our model is diverse.